%% file: main.tex
\documentclass{article}

\usepackage[nonatbib,preprint]{neurips_2024}

\usepackage[utf8]{inputenc} 
\usepackage[T1]{fontenc}    
\usepackage{hyperref}       
\usepackage{url}            
\usepackage{booktabs}       
\usepackage{amsfonts}       
\usepackage{nicefrac}       
\usepackage{microtype}      
\usepackage{xcolor}         

\usepackage{amsmath}
\usepackage{amssymb}
\usepackage{mathtools}
\usepackage{amsthm}

\usepackage{graphicx}
\usepackage{wrapfig}
\usepackage{subcaption}
\usepackage{multirow}
\usepackage{algorithm}
\usepackage{algpseudocode}

\usepackage{diagbox}
\usepackage{hhline}

\title{Adaptive Depth Networks with Skippable Sub-Paths}

\author{%
  Woochul Kang\thanks{Corresponding author} \\
  Department of Embedded Systems\\
  Incheon National University\\
  Yeonsu-gu, Incheon, South Korea, 22012 \\
  \texttt{wchkang@inu.ac.kr} \\
  \And
  Hyungseop Lee \\
  Department of Embedded Systems\\
  Incheon National University\\
  Yeonsu-gu, Incheon, South Korea, 22012 \\
  \texttt{lhhss0927@inu.ac.kr} \\
}

\begin{document}

\maketitle

\input{abstract}    
\input{intro}
\input{related}

\input{method}

\input{experiments}

\input{conc}


\bibliography{main}
\bibliographystyle{nips}

\vfill
\pagebreak

\appendix
\input{appendix}

\end{document}

%% file: abstract.tex
\begin{abstract}
Predictable adaptation of network depths can be an effective way to control inference latency and meet the resource condition of various devices. However, previous adaptive depth networks do not provide general principles and a formal explanation on why and which layers can be skipped, and, hence, their approaches are hard to be generalized and require long and complex training steps. 
In this paper, we present a practical approach to
adaptive depth networks that is applicable 
to both convolutional neural networks (CNNs) and transformers
with minimal training effort. 
In our approach, every hierarchical residual stage is divided into two sub-paths,
and they are trained to acquire different properties
through a simple self-distillation strategy.
While the first sub-path is essential for hierarchical feature learning, the second one is trained to refine the learned features and minimize performance degradation even if it is skipped.
Unlike prior adaptive networks, our approach does not train every target sub-network exhaustively.
At test time, however, we can connect these sub-paths 
in a combinatorial manner 
to select sub-networks of various accuracy-efficiency trade-offs from a single network.
We provide a formal rationale for why the proposed training method can reduce overall prediction errors while minimizing the impact of skipping sub-paths. We demonstrate the generality and effectiveness of our approach with both CNNs and transformers.    
Source codes are available at \href{https://github.com/wchkang/depth}{https://github.com/wchkang/depth}

\end{abstract}

%% file: intro.tex
\section{Introduction}
\label{sec:intro}
Modern deep neural networks
such as CNNs and transformers \cite{vaswani2017attention}
provide state-of-the-art performance
at high computational costs, and,
hence, lots of efforts have been made to leverage those inference capabilities 
in various resource-constrained devices.
Those efforts include 
compact architectures \cite{howard2017mobilenets,han2020ghostnet},
network pruning \cite{compression2016han,liu2019metapruning}, 
weight/activation quantization \cite{jacob2018quantization}, 
knowledge distillation \cite{hinton2015distilling}, to name a few. 
However, those approaches provide static accuracy-efficiency trade-offs,
and, hence, it is infeasible to deploy one single model to meet 
devices with all kinds of resource-constraints. 

There have been some attempts to provide predictable adaptability to neural networks 
by exploiting the redundancy in either 
network depths \cite{huang2018multiscale, fan2020reducing},
widths \cite{yu2018slimmable,alphanet2021wang}, or
both \cite{wan2020orthogonalized,hou2020dynabert}.
However, one major difficulty with prior adaptive networks 
is that they are hard to train and require 
significantly longer training time than non-adaptive networks.
For example, most adaptive networks
select a fixed number of sub-networks
of varying depths or width, 
and train them iteratively,
mostly through self-distilling knowledge from 
the largest sub-network (also referred to as the \emph{super-net}) \cite{yu2018slimmable,alphanet2021wang,hou2020dynabert,touvron2023cotraining}.
However, this exhaustive self-distillation
takes long time and can generate conflicting training objectives
for different parameter-sharing sub-networks, 
potentially resulting in worse performance
\cite{li2019improved,gong2022nasvit}. 
Moreover, unlike width-adaptation networks, 
there is no established principle for adapting depths since the impact of skipping individual layers has not been formally defined.

\begin{figure*}[tbp]
\centering
  \includegraphics[width=1.0\linewidth]{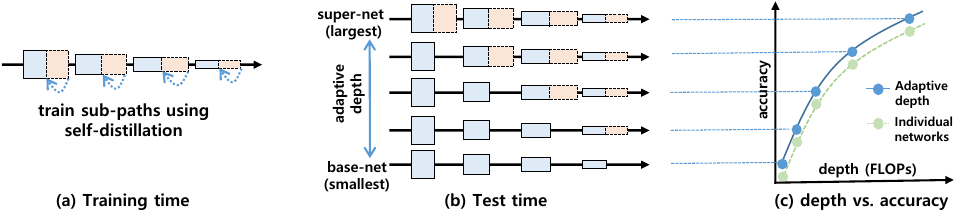}
  \caption{
  (a) During training, every residual stage of 
  a network is divided into two sub-paths.
  The layers in every second (orange) sub-path are optimized to minimize performance degradation even if they are skipped. 
  (b) At test time, these second sub-paths can be skipped in a combinatorial manner, allowing instant selection of various parameter sharing sub-networks. 
  (c) The sub-networks selected from a single network form a better Pareto frontier than counterpart individual networks.
  }
  \label{fig:adaptive-depth-network}
\end{figure*}

In this work, we introduce an architectural pattern and training method for adaptive depth networks that is generally applicable to various networks, e.g., CNNs and transformers. 
In the proposed adaptive depth networks, 
every residual stage is divided into two sub-paths
and the sub-paths are trained to have different properties. 
While the first sub-paths are mandatory for hierarchical feature learning, 
the second sub-paths are optimized to incur minimal performance degradation 
even if they are skipped.  
%
In order to enforce this property of the second sub-paths,
we propose a simple self-distillation strategy,
in which only the largest sub-network (or, \emph{super-net}) 
and the smallest sub-network (or, \emph{base-net}) are
exclusive used as a teacher and a student, respectively,
as shown in Figure-\ref{fig:adaptive-depth-network}-(a).
The proposed self-distillation strategy does not 
require exhaustive training
of every target sub-network, 
resulting in significantly shorter training time than prior adaptive networks.
However, at test time, 
sub-networks with various depths can be selected instantly from a single network 
by connecting these sub-paths in a combinatorial manner, 
as shown in Figure \ref{fig:adaptive-depth-network}-(b). 
Further, these sub-networks with varying depths
outperform individually trained non-adaptive networks 
due to the regularization effect, 
as shown in Figure \ref{fig:adaptive-depth-network}-(c).

In Section \ref{sec:approach},
we discuss the details of our architectural pattern and 
training algorithm,
and show formally that 
the selected sub-paths trained with our self-distillation
strategy are optimized to reduce prediction errors 
while minimally changing the level of input features.
In Section \ref{sec:exp},
we empirically demonstrate that our adaptive depth networks 
outperform 
counterpart individual networks,
both in CNNs and vision transformers,
and achieve actual inference acceleration and energy-saving.


Our approach uniquely introduces a principle for training selected sub-paths to be skippable with minimal performance loss. 
This principle allows us to avoid
typical exhaustive training of target sub-networks 
and instead construct sub-networks
of varying depths from specifically trained sub-paths. 
We anticipate these advances will facilitate 
future research and 
applications of adaptive networks. 

%% file: related.tex
\section{Related Work}
\label{sec:related}

\textbf{Adaptive Networks:} 
In most adaptive networks, parameter-sharing 
sub-networks are instantly selected by adjusting either widths, depths, or resolutions
\cite{huang2018multiscale,yu2018slimmable,alphanet2021wang,hu2019learning,hou2020dynabert, li2019improved,zhang2019your,wan2020orthogonalized,beyer2023flexvit}.
For example, 
slimmable neural networks adjust channel widths of CNN models on the fly 
for accuracy-efficiency trade-offs
and they exploit switchable batch normalization to 
handle multiple sub-networks 
\cite{yu2018slimmable,yu2019universally,alphanet2021wang}.
Transformer-based adaptive depth networks have been proposed for 
language models 
to dynamically skip some of the layers during inference \cite{hou2020dynabert,fan2020reducing}. 
However, in these adaptive networks, 
every target sub-network with varying widths or depths need to be trained 
explicitly, incurring significant training overheads
and potential conflicts between sub-networks.

\textbf{Dynamic Networks:} Dynamic networks \cite{han2021dynamic} are another class of adaptive networks 
that exploit additional control networks or decision gates
for input-dependent adaptation of CNN models  
\cite{wu2018blockdrop, li2021dynamic, guo2019dynamic,li2020learning,yang2020resolution}
and transformers \cite{meng2022adavit,Yin_2022_CVPR,fayyaz2022adaptive,heo2021pit}.
In particular, most dynamic networks
for depth-adaptation have some kinds of  
decision gates at every layer (or block) 
that determine if the layer can be skipped
\cite{figurnov2017spatially,wu2018blockdrop, veit2018convolutional,wang2018skipnet}.
These approaches are based on the thought that some layers
can be skipped on `easy' inputs.
However, the learned policy for skipping layers is opaque to users and 
does not provide a formal description 
of which layers can be skipped for a given input.
Therefore, the network depth cannot be  
controlled by users in a predictable manner
to meet the resource condition of target devices.



\textbf{Residual Blocks with Shortcuts:} 
Since the introduction of ResNets \cite{he2016deep}, 
residual blocks  with identity shortcuts have received extensive attention because of 
their ability to train very deep networks, and have been chosen by
many CNNs \cite{sandler2018mobilenetv2,tan2019efficientnet}
and transformers \cite{vaswani2017attention,dosovitskiy2021an,liu2021Swin}.
Veit et al. \cite{veit2016residual} argue that identity shortcuts make
exponential paths and results in an ensemble of shallower sub-networks. 
This thought is supported by the fact that removing individual residual blocks 
at test time does not significantly affect performance, 
and it has been further exploited to train deep networks 
\cite{huang2016deep, xie2020self}.
Other works argue that identity shortcuts enable residual blocks to perform 
iterative feature refinement, where each block improves slightly but keeps the semantic of the representation of the previous layer
\cite{greff2016highway,jastrzebski2018residual}. 
Our work build upon those views on residual blocks with shortcuts
and further extend them for adaptive depth networks by introducing a training method that enforces the properties of residual blocks more explicitly for skippable sub-paths. 

%% file: method.tex
\section{Adaptive Depth Networks}
\label{sec:approach}

We first present 
the architecture and training details of adaptive depth networks.
Then, we discuss the theoretic rationale for how depth adaptation can be achieved with minimal performance degradation.

\begin{figure*}[tbp!]
\centering
  \includegraphics[width=0.8\linewidth]{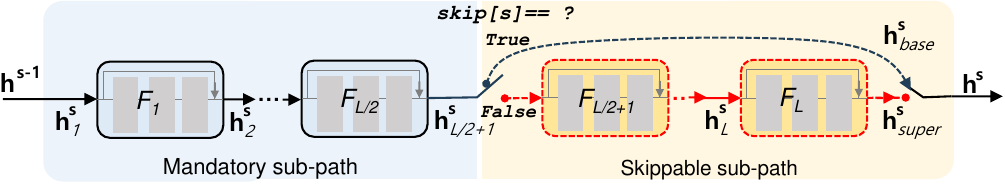}
  \caption{Illustration of 
  a residual stage with two sub-paths.
  While the first (blue) sub-path is mandatory for hierarchical feature learning,
  the second (orange) sub-path can be skipped 
  for efficiency.
  The layers in the skippable sub-path are trained to 
  preserve the feature distribution from $\mathbf{h}^s_{base}$ to $\mathbf{h}^s_{super}$ 
  using the proposed self-distillation strategy.
  Having similar distributions, either
  $\mathbf{h}^s_{base}$ or $\mathbf{h}^s_{super}$ can
  be provided as input $\mathbf{h}^s$ to the next residual stage.
  In the mandatory sub-path, 
  another set of batch normalization operators, called
  \emph{skip-aware BNs}, is exploited
  if the second sub-path is skipped. 
  These sub-paths are building blocks 
  to construct sub-networks of varying depths.
  }
  \label{fig:arch}
\end{figure*}

\subsection{Architectural Pattern for Depth Adaptation}
\label{sec:netarch}
In modern hierarchical networks, such as ResNets \cite{he2016deep} and Swin transformers
\cite{liu2021Swin},
there are typically 4 to 7 consecutive residual stages.\footnote{For vision transformers, residual blocks/stages refer to encoder blocks/stages.
 }
In a network with $N_r$ stages,
the $s$-th ($s=1,...,N_r$) stage consists of $L$ (typically $L \in [3, 6]$) identical residual blocks that  transform input features $\mathbf{h}^{s}_1$ 
additively to produce the output features $\mathbf{h}^{s}$,
as follows:
\begin{align}
 \underbrace{
\mathbf{h}^{s}_{1}}_{=\mathbf{h}^{s-1}} + 
 \underbrace{
 F_{1}(\mathbf{h}^{s}_{1}) +...+{F}^{s}_{L/2}(\mathbf{h}^{s}_{L/2})}_{\mathbf{F}^{s}_{base}}
   + 
  \underbrace{
  F_{L/2+1}(\mathbf{h}^{s}_{L/2+1})... + F_{L}(\mathbf{h}^{s}_{L})}_{\mathbf{F}^{s}_{skippable}}
 = \mathbf{h}^{s}
 \label{eq:residualstage}  
\end{align}
While a block with a residual function $F_\ell$ ($\ell=1,...,L$) 
learns hierarchical features 
as traditional compositional networks \cite{verydeep2015simonyan},
previous literature \cite{jastrzebski2018residual,greff2016highway}
demonstrates that a residual function also 
tend to learn a function that 
refines already learned features at the same feature level.
If a residual block mostly performs feature refinement
while not changing the level of input features, 
the performance of the residual network is not significantly
affected by dropping the block at test time \cite{huang2016deep, xie2020self}.
However, in typical residual networks, most residual blocks 
tend to refine features while learning new level features as well,
and, hence, random dropping of residual blocks at test time 
degrades inference performance significantly.
Therefore, we hypothesize
that if some selected residual blocks can be encouraged explicitly
during training to focus more on feature refinement,
then these blocks can be skipped to save computation 
at marginal loss of prediction accuracy at test time. 

To this end, 
we propose an architectural pattern for adaptive depth networks, in which 
every residual stage is divided into two
sub-paths, or
$\mathbf{F}^{s}_{base}$ and $\mathbf{F}^{s}_{skippable}$ as in Equation \ref{eq:residualstage}
and Figure \ref{fig:arch}.
We train these two sub-paths to have 
different properties (Section \ref{sec:distillation}). 
While $\mathbf{F}^{s}_{base}$ is trained to learn 
feature representation $\mathbf{h}^s_{base}$  (= $\mathbf{h}^s_{ L/2+1}$) with no constraint, 
the second sub-path $\mathbf{F}^{s}_{skippable}$ is 
constrained to
preserve the feature level of $\mathbf{h}^s_{base}$ 
and only refine it to produce $\mathbf{h}^s_{super} (= \mathbf{h}^s_{L+1}$).
Since layers in $\mathbf{F}^{s}_{base}$ perform
essential transformations for hierarchical feature learning, they cannot be bypassed during inference.
In contrast, layers in $\mathbf{F}^{s}_{skippable}$ can be skipped 
for efficiency since they only refine $\mathbf{h}^s_{base}$.
If $\mathbf{F}^{s}_{skippable}$ is skipped, then
$\mathbf{h}^{s}_{base}$ becomes the input to the next residual stage. 
Therefore, $2^{N_r}$ (=$\sum_{k=0}^{N_r}{C(N_r, k)}$) sub-networks 
with varying accuracy-efficiency trade-offs
can be selected from a single network
by choosing whether to skip $\mathbf{F}^{s}_{skippable} (s=1,...,N_r)$ or not
(Table \ref{tab:subnet-perf}).

This architectural pattern is agnostic to the type of residual blocks.
The residual function $F$ can be convolution layers for CNNs
and self-attention + MLP layers for transformers.
Table \ref{tab:arch-details} in Appendix \ref{sec:arch-detail} provides details on how this pattern is applied to both CNNs and transformers. 


\subsection{Training Sub-Paths with Self-Distillation}
\label{sec:distillation}

Preserving the feature level of $\mathbf{h}^{s}_{base}$
in $\mathbf{F}^{s}_{skippable}$
implies, more specifically, that 
two feature representations $\mathbf{h}^{s}_{base}$ and 
$\mathbf{h}^{s}_{super}$ have similar distributions over training input $\mathbf{X}$.
Algorithm \ref{alg:training} shows 
our training method, in which 
$\mathbf{h}^{s}_{base}$ and 
$\mathbf{h}^{s}_{super}$ are encouraged to have similar distributions
by including 
\emph{Kullback-Leibler} (KL) divergence between them,
or ${D_{KL}}(\mathbf{h}^s_{super} || \mathbf{h}^s_{base})$,
in the loss function. 

\begin{algorithm}[tbh!]
\caption{Training algorithm for an adaptive depth network $\mathbf{M}$. 
The forward function of $\mathbf{M}$ accepts an extra argument, `\textit{skip}', 
which controls the residual stages where their skippable sub-paths are skipped.
For example,
the smallest sub-network, or \emph{base-net}, of $\mathbf{M}$ is selected by passing `\textit{skip=[True, True, True, True]}'
when the total number of residual stages,
denoted by $N_r$, is 4.}
\label{alg:training}
\begin{algorithmic}[1]
\State Initialize an adaptive depth network $\mathbf{M}$
\For{$i=1$ {\bfseries to} $n_{iters}$}
\State  Get next mini-batch of data $\mathbf{x}$ and label $\mathbf{y}$
\State  $optimizer.zero\_grad()$ 
\State $\mathbf{\hat{y}}_{super}, \mathbf{h}_{super} = \mathbf{M}.forward(\mathbf{x}, \textrm{skip=[False, ..., False]})$ 
\Comment{forward pass for \emph{super-net}}
\State $loss_{super}= criterion(\mathbf{y}, \mathbf{\hat{y}}_{super})$
\State $loss_{super}.backward()$ 
\State $\mathbf{\hat{y}}_{base}, \mathbf{h}_{base} = \mathbf{M}.forward(\mathbf{x}, \textrm{skip=[True, ..., True]})$ 
\Comment{forward pass for \emph{base-net}}
\State $loss_{base}=\sum_{s=1}^{N_{r}} D_{KL}(\mathbf{h}^s_{super} \| \mathbf{h}^s_{base}) 
+ D_{KL}(\mathbf{\hat{y}}_{super} \| \mathbf{\hat{y}}_{base})$
\State $loss_{base}.backward()$
\Comment{self-distillation of skippable sub-paths}
\State $optimizer.step()$
\EndFor
\end{algorithmic}
\end{algorithm}
In Algorithm \ref{alg:training}, the largest and the smallest sub-networks of $\mathbf{M}$, 
which are called \emph{super-net} and the \emph{base-net}, respectively, are exploited.
In steps 9-10, 
the hierarchical features (and outputs) from the super-net and the base-net are encouraged to have similar distributions through 
self-distillation. 
Due to the architectural pattern of
interleaving the mandatory and the skippable sub-paths, 
minimizing $D_{KL}(\mathbf{\hat{y}}_{super} \| \mathbf{\hat{y}}_{base})$ also
minimizes $D_{KL}(\mathbf{h}_{super} \| \mathbf{h}_{base})$ implicitly. 
Therefore, only $D_{KL}(\mathbf{\hat{y}}_{super} \| \mathbf{\hat{y}}_{base})$ can
be included in $loss_{base}$ if the extraction of intermediate features is tricky.




Self-distillation has been extensively used in prior adaptive networks \cite{alphanet2021wang, yu2018slimmable, hou2020dynabert}. 
However, their primary goal is to train every target sub-network. 
For example, on each iteration, sub-networks are randomly sampled from a large search space to act as either teachers or students \cite{alphanet2021wang},
which takes significantly longer training time
and can generate conflicting training objectives
for different parameter-sharing sub-networks \cite{li2019improved,gong2022nasvit}. 
In contrast, our method in Algorithm 1 
focuses on training sub-paths
by exclusively using the super-net as a teacher and the base-net as a student. 
By focusing on sub-paths, 
the training procedure in Algorithm \ref{alg:training} is significantly simplified
and avoids potential conflicts among sub-networks.
At test time, however,  
various sub-networks
can be selected by connecting these sub-paths.
The effect of this self-distillation strategy
is investigated in Section \ref{sec:ablation}.

\subsection{Analysis of Skippable Sub-Paths}
\label{sec:theory}
\textbf{Formal Analysis}:
$D_{KL}(\mathbf{h}^s_{super} || \mathbf{h}^s_{base})$ in the loss function 
$loss_{base}$
can be trivially minimized if residual blocks in 
$\mathbf{F}^{s}_{skippable}$ learn identity functions,
or $\mathbf{h}^s_{base} + \mathbf{F}^{s}_{skippable}(\mathbf{h}^s_{base}) = \mathbf{h}^s_{base}$.
However, since the super-net is jointly trained 
with the loss function $loss_{super}$, 
the residual functions in $\mathbf{F}^{s}_{skippable}$ cannot simply be an identity function. 
Then, what do the residual functions in $\mathbf{F}^{s}_{skippable}$ learn during training?
This can be further investigated through Taylor expansion
\cite{jastrzebski2018residual}.
For our adaptive depth networks, 
a loss function $\mathcal{L}$ used for training
the super-net 
can be approximated with Taylor expansion as follows:
\begin{align} 
 & \mathcal{L}(\mathbf{h}^{s}_{super})  
 = \mathcal{L} \{\mathbf{h}^s_{base} + \mathbf{F}^{s}_{skippable}(\mathbf{h}^s_{base})\} \\
 & =\mathcal{L}\{\mathbf{h}^s_{base} + F_{L/2+1}(\mathbf{h}^{s}_{L/2+1}) + ... + F_{L-1}(\mathbf{h}^{s}_{L-1})+ F_{L}(\mathbf{h}^{s}_{L})\}
 \label{eq:taylorfirst} \\
 & \approx \mathcal{L}\{\mathbf{h}^s_{base} + F_{L/2+1}(\mathbf{h}^{s}_{L/2+1}) + ... + F_{L-1}(\mathbf{h}^{s}_{L-1})\} 
 + F_{L}(\mathbf{h}^{s}_{L})\cdot \frac{\partial \mathcal{L}(\mathbf{h}^s_{L})}{\partial \mathbf{h}^s_{L}} + \mathcal{O}(F_L(\mathbf{h}_{L}^s)) \label{eq:taylormiddle} 
\end{align}

In Equation \ref{eq:taylormiddle}, the loss function is expanded around $\mathbf{h}^s_{L}$, or $ \mathbf{h}^s_{base} + ... + F_{L-1}(\mathbf{h}^{s}_{L-1})$.
Only the first order term 
is left and all high order terms, such as 
$ F_L(\mathbf{h}_{L}^s)^2 \cdot \frac{ {\partial^2\mathcal{L}(\mathbf{h}^s_{L})} } { 2 {\partial(\mathbf{h}^s_{L})^2} }$, 
are absorbed in $\mathcal{O}(F_L(\mathbf{h}_{L}^s))$.

\begin{wrapfigure}{r}{0.46\linewidth}
\begin{center}
\vspace{-0.15in}
\includegraphics[width=1.0\linewidth]{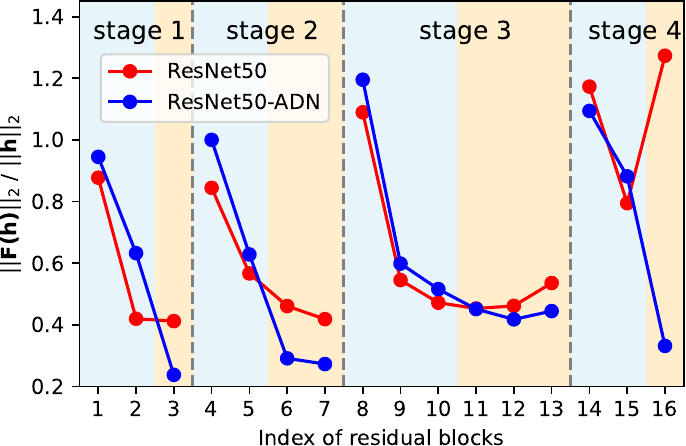}
  \caption{$||F(\mathbf{h})||_2 / ||\mathbf{h}||_2$ at residual blocks.
  In ours, skippable sub-paths (orange areas) minimally change the distribution of input $\textbf{h}$.}
  \label{fig:l2-subpaths}
  \vspace{-0.1in}
\end{center}
\end{wrapfigure}

The high-order terms in $\mathcal{O}(F_L(\mathbf{h}_{L}^s))$ can be ignored
if $F_L(\mathbf{h}_{L}^s)$ has a small magnitude.
In typical residual networks, however,
every layer is trained to learn new features 
with no constraint, and, hence, 
there is no guarantee that $F_L(\mathbf{h}^s_L)$
have small magnitude.
In contrast, in our adaptive depth networks,
the residuals 
in $\mathbf{F}^{s}_{skippable}$ 
are explicitly enforced to have small magnitude through 
the proposed self-distillation strategy
(refer to Figure \ref{fig:l2-subpaths} for empirical evidence).
As a result, the terms in $\mathcal{O}(F_L(\mathbf{h}_{L}^s))$
can be ignored for the approximation.
If we similarly keep expanding the loss function around $\mathbf{h}^s_j$ ($j= L/2 +1,...,L$)
while ignoring high order terms,
we obtain the following approximation:
 \begin{align}
 \mathcal{L}(\mathbf{h}^{s}_{super}) \approx \mathcal{L}(\mathbf{h}^{s}_{base}) + \sum_{j={L/2+1}}^{L} F_{j}(\mathbf{h}^s_{j}) \cdot 
 \frac{\partial \mathcal{L}(\mathbf{h}^s_{j})}{\partial \mathbf{h}^s_{j}} \label{eq:taylorfinal}
\end{align}

In Equation \ref{eq:taylorfinal}, 
minimizing the loss $\mathcal{L}(\mathbf{h}^s_{super})$
during training drives $F_{j}(\mathbf{h}^s_{j})$ $(j=L/2+1,...,L)$ 
in the negative half space of
$\textrm{ }\frac{\partial \mathcal{L}(\mathbf{h}^s_{j})}{\partial \mathbf{h}^s_{j}}$
to minimize the dot product between $F_{j}(\mathbf{h}^s_{j})$ and $\frac{\partial \mathcal{L}(\mathbf{h}^s_{j})}{\partial \mathbf{h}^s_{j}}$.
This implies that every residual function in $\mathbf{F}^{s}_{skippable}$
is optimized to learn a function 
that has a similar effect to gradient descent:
\begin{align}
F_{j}(\mathbf{h}^{s}_{j}) \simeq \mathbf{-} \textrm{ }\frac{\partial \mathcal{L}(\mathbf{h}^s_{j})}{\partial \mathbf{h}^s_{j}} \textrm{ }(j=L/2+1,...,L)
\label{eq:residualnegativehalf}
\end{align}
In other words, the residual functions in the skippable sub-paths reduce 
the loss $\mathcal{L}(\mathbf{h}^s_{base})$ iteratively during inference
while preserving the feature distribution of $\mathbf{h}^s_{base}$.

Considering this result,
we can conjecture that,
with our architectural pattern and self-distillation strategy, 
layers in $\mathbf{F}^{s}_{skippable}$ 
learn functions that refine 
input features $\mathbf{h}^s_{base}$
iteratively for better inference accuracy
while minimally changing the distribution of $\mathbf{h}^s_{base}$.

\textbf{Empirical Analysis}: 
We can estimate how much the distribution of input $\mathbf{h}$
is transformed by residual function $F$ 
by measuring $||F(\mathbf{h})||_2 / ||\mathbf{h}||_2$
at each residual block.
Figure \ref{fig:l2-subpaths} illustrates $||F(\mathbf{h})||_2 / ||\mathbf{h}||_2$ at every residual block in the baseline ResNet50
and our ResNet50-ADN that is trained according to Algorithm \ref{alg:training}.\footnote{ImageNet validation dataset
is used for this experiment.}
Our ResNet50-ADN exhibits greater transformation than ResNet50 in the mandatory sub-paths (blue areas)
and less transformation in the skippable sub-paths (orange areas).
This result demonstrates that our self-distillation strategy in Algorithm \ref{alg:training} effectively trains the skippable sub-paths to minimally change the 
input distribution. 
As a result, the blocks in skippable sub-paths 
can be skipped with less impact on performance. 

\subsection{Skip-Aware Batch/Layer Normalization}
\label{sec:batchnorm}

Originally, batch normalization (BN) \cite{ioffe2015batch} 
was proposed to handle internal covariate shift
during training non-adaptive networks by normalizing features.
In our adaptive depth networks, however, 
internal covariate shifts can occur
during inference
in mandatory sub-paths if different sub-networks are selected.
To handle potential internal covariate shifts, 
switchable BN operators, called \emph{skip-aware BNs},
are used in mandatory sub-paths.
For example, at each residual stage,
two sets of BNs are available for the mandatory sub-path,
and they are switched
depending on whether its skippable sub-path is skipped or not.

The effectiveness of switchable BNs has been demonstrated 
in networks with adaptive widths \cite{yu2019autoslim, alphanet2021wang}
and adaptive resolutions \cite{mingian2021dynamic}.
However, in previous adaptive networks, 
$N$ sets of switchable BNs are required in every layer
to support $N$ parameter-sharing sub-networks.
Such a large number of switchable BNs not only 
requires more parameters,
but also makes the training process complicated 
since $N$ sets of switchable BNs
need to be trained iteratively during training. 
In contrast, in our adaptive depth networks, 
every mandatory sub-path needs only two sets of switchable BNs,
regardless of the number of supported sub-networks.
This reduced number of switchable BNs 
significantly simplifies the 
training process as shown in Algorithm \ref{alg:training}.
Furthermore, 
the amount of parameters for skip-aware BNs is negligible.
For instance, in ResNet50, skip-aware BNs increase the parameters by 0.07\%. 

Transformers \cite{vaswani2017attention,liu2021Swin} exploit layer normalization (LN) 
instead of BNs and naive replacement of LNs to BNs incurs instability during training \cite{yao2021leveraging}.
Therefore, for our adaptive depth transformers,
we apply switchable LN operators in mandatory sub-paths instead of switchable BNs.

%% file: experiments.tex
\section{Experiments}
\label{sec:exp}



We use 
networks both from CNNs and vision transformers 
as base models to apply the proposed architecture pattern:
MobileNet V2 \cite{sandler2018mobilenetv2} is a lightweight CNN model,
ResNet \cite{he2016deep} is a larger CNN model,
and 
ViT \cite{dosovitskiy2021an} and Swin-T \cite{liu2021Swin}
are representative vision transformers.
All base models except ViT have hierarchical
stages, each with 2 $\sim$ 6 residual blocks.
So, 
according to the proposed architectural pattern,
every stage is evenly divided into 2 sub-paths
for depth adaptation.
ViT does not define hierarchical stages
and all residual encoder blocks have same 
spatial dimensions.
Therefore, we divide 12 encoder blocks into 4 groups,
resembling other residual networks, and 
select the last encoder block of each group 
as a skippable sub-path.
Details are in Table \ref{tab:arch-details} in Appendix \ref{sec:arch-detail}.
These models are trained according to Algorithm \ref{alg:training}. 
For self-distillation, only the final outputs from the super-net and the base-net,
or $\mathbf{\hat{y}}_{super}$ and $\mathbf{\hat{y}}_{base}$ respectively, are 
used since exploiting intermediate features for explicit self-distillation  
has a marginal impact on performance (Section \ref{sec:ablation}). 

We use the suffix \emph{`-ADN'} to denote our adaptive depth networks.
A series of boolean values in parentheses denotes a specific sub-network used for evaluation; 
each boolean value represents the residual stage where its skippable sub-path is skipped. 
For example, ResNet50-ADN (FFFF) and ResNet50-ADN (TTTT) correspond to the super-net and the base-net of ResNet50-ADN, respectively.



\renewcommand{\dblfloatpagefraction}{0.8}
\renewcommand{\arraystretch}{1.1}

\begin{figure}[tbp!]
\centering
\begin{footnotesize}
  \subfloat[Results on ImageNet]{
  \renewcommand{\arraystretch}{0.8}
  \begin{tabular}[tbp]{p{3.2cm}p{0.8cm}rrr}
    \toprule
    \multirow{2}{*}{Model}     & Params     & FLOPs & Acc \\ 
         & (M)     & (G) & (\%) \\ 
    \midrule
    \textbf{ResNet50-ADN (FFFF)} & \multirow{2}{*}{25.58} & 4.11 & 77.6 \\ 
    \textbf{ResNet50-ADN (TTTT)} &  & 2.58 & 76.1 \\ 
    ResNet50  &  25.56  & 4.11 & 76.7 \\ 
    ResNet50-Base   & 17.11 & 2.58 & 75.0 \\
    \midrule
    \textbf{MbV2-ADN (FFFFF)} & \multirow{2}{*}{3.53} & 0.32 & 72.5 \\
    \textbf{MbV2-ADN (TTTTT)} &  &  0.22 & 70.6 \\ 
    MbV2 &  3.50  & 0.32 & 72.1 \\ 
    MbV2-Base   & 2.98 & 0.22 & 70.2 \\
    
    \midrule
    \textbf{ViT-b/16-ADN (FFFF)} & \multirow{2}{*}{86.59} & 17.58 & 81.4 \\
    \textbf{ViT-b/16-ADN (TTTT)} &  & 11.76  & 80.6 \\
    ViT-b/16  &  86.57  & 17.58 & 81.1  \\ 
    ViT-b/16-Base   & 67.70 & 11.76 & 78.7 \\
    \midrule
    \textbf{Swin-T-ADN (FFFF)} & \multirow{2}{*}{28.30} & 4.49 & 81.6 \\ 
    \textbf{Swin-T-ADN (TTTT)} &  & 2.34  & 78.0 \\ 
    Swin-T  &  28.29  & 4.49 &  81.5 \\ 
    Swin-T-Base   & 15.34 & 2.34 & 77.4 \\
   

    \bottomrule
  \end{tabular}
  \renewcommand{\arraystretch}{1.0}
  \label{table:ilsvrceval}
}
\end{footnotesize}
  \subfloat[Pareto frontiers of our networks]{
  \label{fig:pareto-front}
  \includegraphics[width=0.43\linewidth]{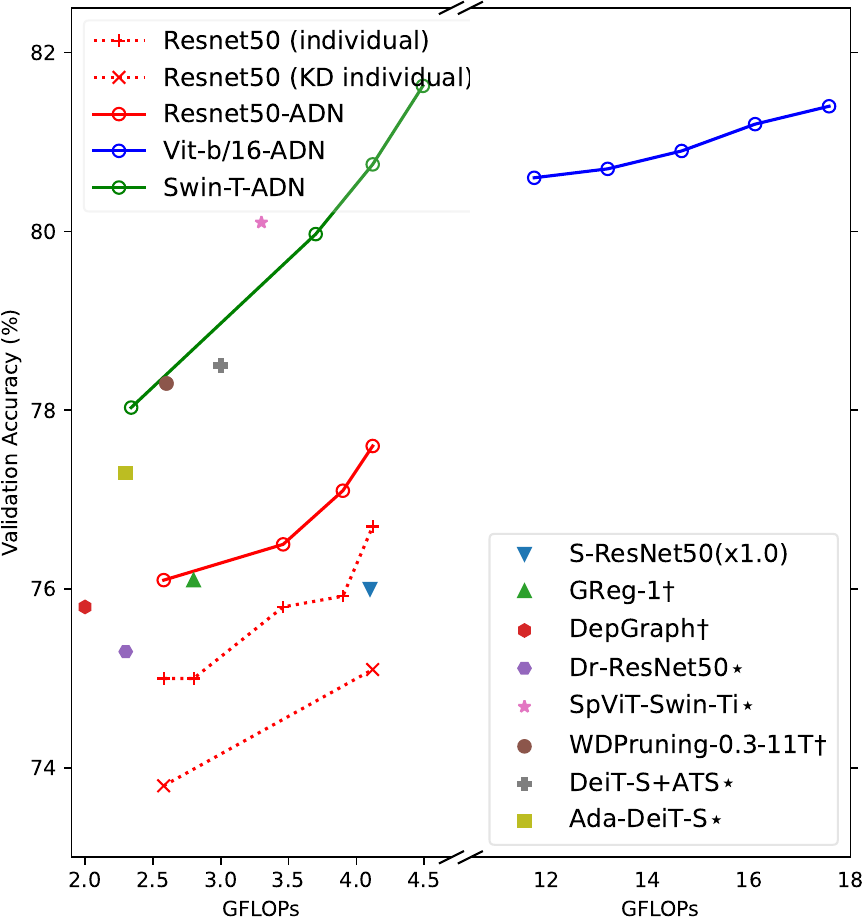}
}
\caption{
  (a) Results on ImageNet validation dataset. Networks with the suffix `-Base' have the same depths as
  the base-nets of corresponding adaptive depth networks.
  (b) Pareto frontiers formed by the sub-networks of our adaptive depth networks.
  ResNet50 (individual) and ResNet50 (KD individual) are non-adaptive networks having same depths
  as the sub-networks of ResNet50-ADN.
}
\label{fig:exp-ilsvrc}
\end{figure}
\renewcommand{\arraystretch}{1.0}

\subsection{ImageNet Classification}
\label{sec:exp-imagenet}
We evaluate our method on ILSVRC2012 dataset \cite{russakovsky2015imagenet} that has 1000 classes. 
The dataset consists of 1.28M training and 50K validation images. 
For CNN models, we follow most training settings in the original papers
\cite{he2016deep,sandler2018mobilenetv2},
except that ResNet models are trained for 150 epochs.
ViT and Swin-T are trained for 300 epochs,
following DeiT's training recipe \cite{hugo2020training,pytorch2023training}.
For Swin-T-ADN, we disable stochastic depths \cite{huang2016deep} for the 
mandatory sub-paths 
since the strategy of random dropping of residual blocks
conflicts with our approach to skipping sub-paths.
For fair comparison, 
our adaptive depth networks and corresponding individual networks 
are trained in the same training settings.

The results in Figure \ref{fig:exp-ilsvrc}-(a)
show that our adaptive depth networks
outperform counterpart individual networks 
even though many sub-networks share parameters in a single model. 
Further, 
our results with vision transformers demonstrate that 
our approach is generally applicable 
and compatible with their state-of-the-art
training techniques such as 
DeiT's training recipe \cite{hugo2020training}.
We conjecture that this performance improvement 
results from effective distillation of knowledge from $\textbf{h}^s_{super}$
to $\textbf{h}^s_{base}$ at each residual stage
and the iterative feature refinement
at skippable sub-paths,
shown in Equation \ref{eq:residualnegativehalf}.


Figure \ref{fig:exp-ilsvrc}-(b)
shows Pareto frontiers formed by selected
sub-networks of our adaptive depth networks;
Table \ref{tab:subnet-perf} in Appendix \ref{sec:subnet-perf} shows the performance of all sub-networks.
In Figure \ref{fig:exp-ilsvrc}-(b) and Table \ref{table:imagenet-sota},
several state-of-the-art efficient inference methods
and dynamic networks 
are compared with our base-networks.
The result demonstrates that 
our adaptive depth networks show comparable performance 
across a range of FLOPs.
In Figure \ref{fig:exp-ilsvrc}-(b),
it should be noted that individual
ResNets trained with knowledge distillation
has worse performance than individual ResNets. 
As reported in previous works,
successful knowledge distillation requires 
a patient and long training \cite{beyer2022knowledge},
and straightforward knowledge distillation using ImageNet
does not improve the performance of student models \cite{payingmore2017zagoruyko, cho2019efficacy}.
In contrast, our ResNet50-ADN trained with 
the proposed self-distillation strategy achieves better performance 
than counterpart ResNets.
This demonstrates that the high performance of 
adaptive depth networks does not simply 
come from distillation effect.

\begin{table*}[tbp!]
\centering
\begin{footnotesize}
\caption{Our base-nets are compared with state-of-the-art efficient inference methods.
  $\dagger$ denotes static pruning methods,
  $\ast$ denotes width-adaptation networks,
  and $\star$ denotes input-dependent dynamic networks.
  While these approaches exploit various non-canonical training techniques, such as iterative retraining,
  our base-nets are instantly selected from adaptive depth networks without fine-tuning. 
}
\renewcommand{\arraystretch}{0.8}
  \begin{tabular}[tbh!]{llccc}
    \toprule
    Model      & FLOPs & $\downarrow$FLOPs & Acc@1 \\
    \midrule
    GReg-1 \cite{wang2021neural}$\dagger$  & 2.8G & 33\% & 76.1\% \\
    DepGraph \cite{fang2023depgraph}$\dagger$  & 2.0G & 51\% & 75.8\%  \\
    DR-ResNet50 ($\alpha$=2.0) \cite{mingian2021dynamic}$\star$ &  2.3G & 44\% & 75.3\% \\  
    \textbf{ResNet50-ADN (TTTT)} &  2.6G & 37\% & 76.1\%  \\  
    \midrule
    AlphaNet-0.75x \cite{alphanet2021wang} $\ast$ & 0.21G & 34\% & 70.5\% \\
    \textbf{MbV2-ADN (TTTTT)} & 0.22G & 32\% & 70.6\%  \\ 
    
    \midrule
    WDPruning-0.3-11 \cite{yu2022widthdepth}$\dagger$ & 2.6G & - & 78.3\% \\
    X-Pruner \cite{yu2023xpruner} $\dagger$ & 3.2G &  28\%  & 80.7\% \\
    Ada-DeiT-S \cite{meng2022adavit}$\star$ & 2.3G &  - & 77.3\%  \\
    SPViT-Swin-Ti \cite{kong2022spvit}$\star$ & 3.3G &  27\% & 80.1\% \\
    \textbf{Swin-T-ADN (TTTT)} &  2.3G & 48\% & 78.0\%  \\ 

    \bottomrule
  \end{tabular}
\renewcommand{\arraystretch}{1.0}
\label{table:imagenet-sota}
\end{footnotesize}
\end{table*}
\renewcommand{\arraystretch}{1.2}

\subsection{Training Time and Sub-Networks}
\label{sec:training-time}
\begin{figure}[tbp!]
\centering
\begin{footnotesize}
\subfloat[Training time]{
  \renewcommand{\arraystretch}{0.9}
  \begin{tabular}[htbp!]{c c c}
  \hline 
  Model   & 1 epoch (min) & \# of sub-nets \\
  \hline
  ResNet50 & 32.9 & - \\
  ResNet50-Base  &  22.7 & - \\
  \textbf{ResNet50-ADN (ours)} & 54.7 & $2^4$ depths \\ 
  MSDNet \cite{huang2018multiscale} &  67.1 & 9 depths\\ 
  S-ResNet50 \cite{yu2018slimmable} &  82.5 & 4 widths\\
  \hline
  MbV2  & 13.1 & - \\
  MbV2-Base & 10.0 & - \\
  \textbf{MbV2-ADN (ours)} & 22.5 & $2^5$ depths \\
  AlphaNet$^*$ \cite{alphanet2021wang} &  230.5 & 216 depths \\  
  \hline
  \end{tabular}
  \renewcommand{\arraystretch}{1.0}
  \label{tab:trainingtime}
}
\quad
  \subfloat[On-device Inference latency/energy]
  {
    \includegraphics[width=0.37\linewidth]{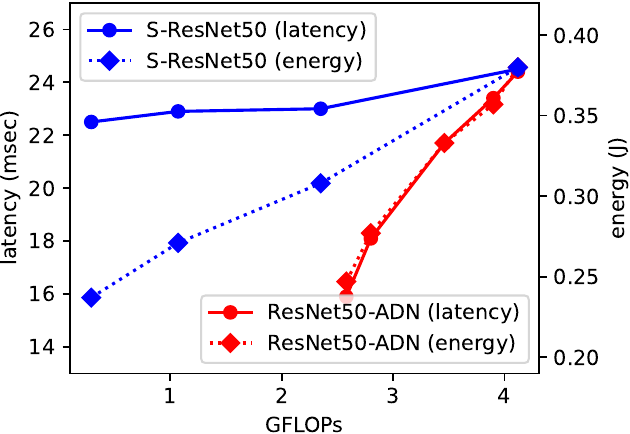}
    \label{fig:flops.vs.latency}
  }
\caption{
(a) Training time (1 epoch),
measured on Nvidia RTX 4090 (batch size: 128).
AlphaNet$^*$ is configured to have similar FLOPs
to MbV2 and makes sub-networks by only adjusting the network depth.
(b) Inference latency and energy consumption
  of adaptive networks, measured on Nvidia Jetson Orin Nano (batch size: 1)
}
\label{fig:exp-train-test}
\end{footnotesize}
\end{figure}

One of the key advantages of our training method in Algorithm \ref{alg:training}
is its simplicity and significantly reduced training effort 
compared to previous adaptive networks.
%
%
Figure \ref{fig:exp-train-test}-(a) shows that 
the training time of our adaptive depth networks
are similar to the training time of two individual networks combined.
This is because our training method
trains sub-paths, rather than sub-networks, by exploiting only the super-net and the base-net.
In contrast, the compared adaptive networks 
require much longer training time than ours 
since they have to explicitly 
apply self-distillation to all target sub-networks.
For example, on every training iteration,
AlphaNet \cite{alphanet2021wang} randomly samples sub-networks from 
its search space for self-distillation.
Although the prior works we compared
may seem outdated, they remain relevant and representative because there has been little progress in improving the training cost of adaptive networks.

While AlphaNet \cite{alphanet2021wang} supports
significantly more sub-networks, 
it should be noted that 
supporting many sub-networks is not our goal.
Instead, our objective is 
to provide better performance 
Pareto with several useful sub-networks,
as demonstrated in Figure \ref{fig:exp-ilsvrc}-(b).
For example, 
although 
ResNet50-ADN has $2^4$ sub-networks in its search space,
Table \ref{tab:subnet-perf} in Appendix \ref{sec:subnet-perf}
demonstrates that some sub-networks 
manifest better performance 
than the others even if they have similar depths.
The results 
show that skipping sub-paths in deeper residual stages generally has a more detrimental impact on performance, 
suggesting that feature refinement is more crucial for deeper residual stages.

\subsection{On-Device Performance}
\label{sec:on-device-perf}
While the Pareto frontiers formed by sub-networks
demonstrate theoretic performance,
inference acceleration in actual devices
is more important in practice 
for effective control of inference latency and energy
consumption.

Figure \ref{fig:exp-train-test}-(b) shows the performance 
on Nvidia Jetson Orin Nano.
The inference latency and energy consumption
of  ResNet50-ADN is compared to
S-ResNet50, a representative width-adaptation network. 
The result shows that depth-adaptation of ResNet50-ADN 
is highly effective in accelerating inference speeds 
and reducing energy consumption.
Although our ResNet50-ADN has a limited FLOPs adaptation range, 
reducing FLOPs by 38\% through depth adaptation 
reduces both inference latency and energy consumption by 35\%.
In contrast, even though S-ResNet50 can reduce FLOPs by up to 93\% by adjusting its width, it only achieves up to 9\% acceleration in practice.

\subsection{Ablation Study}
\label{sec:ablation}

\begin{table*}[tbhp!]
\centering
\begin{footnotesize}
\caption{Ablation analysis with ResNet50-ADN and ViT-b/32-ADN.
  Applied components are checked. 
  $\downarrow$ and $\uparrow$ in parentheses 
  are comparisons to non-adaptive individual networks.
  By default, only the outputs, or $\mathbf{\hat{y}}_{super}$ and $\mathbf{\hat{y}}_{base}$, are used for self-distillation.
  The last row with double check marks shows the results when both 
  intermediate features and outputs are used for self-distillation. 
  }
  \label{tab:ablation}
  \begin{tabular}[tbhp]{cc|cc|cc}
  \hline 
  self-  & skip-aware & \multicolumn{2}{c|}{ResNet50-ADN Acc@1 (\%)} & \multicolumn{2}{c}{ViT-b/32-ADN Acc@1 (\%)} \\
  \cline{3-6} 
  distllation. &  BNs/LNs & FFFF & TTTT & FFFF & TTTT \\
  \hline
    &    & 75.2\% ($\downarrow$ 1.5\%) & 72.2\% ($\downarrow$ 2.8\%) & 75.7\% ($\downarrow$0.2\%) & 74.1\% ($\uparrow$ 0.3\%)\\
  \hline
  \checkmark   &  & 76.1\% ($\downarrow$ 0.6\%) & 74.9 \% ($\downarrow$ 0.1\%) & 76.4\% ($\uparrow$0.5\%) & 74.3\% ($\uparrow$0.5\%)\\
   \hline
    &  \checkmark &  76.6\% ($\downarrow$ 0.1\%) & 75.1\% ($\uparrow$ 0.1\%) &  76.0\% ($\uparrow$0.1\%) & 74.3\% ($\uparrow$0.5\%)\\
   \hline
    \checkmark & \checkmark  & 77.6\% ($\uparrow$ 0.9\%) & 76.1\% ($\uparrow$ 1.1\%) & 76.6\% ($\uparrow$ 0.8\%) & 74.3\%  ($\uparrow$0.5\%)\\
  \hline
    \checkmark\checkmark & \checkmark  & 77.3\% ($\uparrow$ 0.6\%) & 76.2\% ($\uparrow$ 1.2\%) &  & \\
   \hline
  \end{tabular}
\end{footnotesize}
\end{table*}

We first investigate
the influence of two key components of the proposed adaptive depth networks:
(1) self-distillation of sub-paths and (2) skip-aware BNs/LNs.
When our self-distillation method is not applied, 
the loss of the base-net, or $loss_{base}$, in Algorithm \ref{alg:training} is modified to 
$criterion(\mathbf{y}, \mathbf{\hat{y}}_{base})$.
Table \ref{tab:ablation} shows the results.
For ResNet50-ADN, when neither of them is applied, the inference accuracy of the super-net 
and the base-net is significantly lower than 
non-adaptive individual networks by 1.5\% and 2.8\%, respectively.
This result shows the difficulty of joint training 
sub-networks for adaptive networks. 
When one of the two components is applied individually,
the performance is still slightly worse than individual networks'.
When both self-distillation and skip-aware BNs are applied together, 
ResNet50-ADN achieves significantly better performance than 
individual networks, both in the super-net and the base-net.
The last row, with double check marks, shows that
exploiting intermediate features as well as softmax outputs
for self-distillation has minor impact on performance. 

\begin{table}[h!]
    \centering
    \begin{footnotesize}
         \caption{Comparison of self-distillation strategies. 
         Our approach (in bold) uses exclusively 
         the super-net and the base-net
         as a teacher and a student, respectively.
         }
         \label{tab:abs-study-random}
    \begin{tabular}{ p{0.9cm} p{0.9cm} | p{0.6cm} p{0.6cm} p{0.6cm} p{0.6cm} p{0.6cm} | p{0.6cm} p{0.6cm} p{0.6cm} p{0.6cm} p{0.6cm} }
     \multicolumn{2}{c|}{\diagbox{}{sub-nets }} & \multicolumn{5}{c|}{ResNet50-ADN Acc@1 (\%)} 
        & \multicolumn{5}{c}{ViT-b/32-ADN Acc@1 (\%)} \\    
     \cline{3-12}
     Teacher & Student   & FFFF & TFFF & TTFF & TTTF & TTTT 
        & FFFF & TFFF & TTFF & TTTF & TTTT \\    
    \hline \hline
    \textbf{FFFF} & \textbf{TTTT} & \textbf{77.6} & \textbf{77.1} & \textbf{76.5} & \textbf{76.0} & \textbf{76.1} &
        \textbf{76.6} & \textbf{76.0} & \textbf{75.5} & \textbf{74.5} & \textbf{74.3} \\
    \hline
    FFFF & Random & 77.1 & 76.7 & 76.4 & 75.5 & 74.8 &
        75.2 & 74.7 & 73.9 & 72.9 & 71.1 \\ 
    \hline
    Random & TTTT & 75.5 & 75.5 & 75.4 & 75.0 & 74.9 &
        72.0 & 72.0 & 71.9 & 71.8 & 71.7 \\ 
    \hline
    Random & Random & 75.4 & 75.2 & 75.2 & 74.9 & 74.6 &
        70.8 & 70.7 & 70.6 & 70.4 & 70.3 \\
    \hline
   \end{tabular}
   \end{footnotesize}
\end{table}
\textbf{Self-Distillation Strategies}: Our self-distillation approach
in Algorithm \ref{alg:training}
exploits only two sub-networks exclusively as a teacher and a student.
Specifically, the super-nets, or FFFF, acts as the the teacher and the base-nets, or TTTT, becomes the student. 
The purpose of this strategy is not to train only those two sub-networks, but rather to train skippable sub-paths in a way that minimally modifies the feature distribution, as demonstrated in Figure \ref{fig:l2-subpaths}. 
To investigate the effect of self-distillation strategies, 
we conduct an experiment in Table \ref{tab:abs-study-random}. 
In every training iteration, rather than exclusively using FFFF and TTTT sub-networks for self-distillation, 
we randomly sample either a teacher, a student, or both from $2^4$ sub-networks. 
These randomly sampled sub-networks are trained explicitly through self-distillation.
However, as shown in Table \ref{tab:abs-study-random}, 
all sub-networks, both from ResNet50-ADN and ViT-b/32-ADN, perform significantly better when our self-distillation strategy is applied for training.
Even though most of our sub-networks (such as TFFF, TTFF, and TTTF) are
instantly constructed at test time without explicit training, they still outperform their counterpart sub-networks trained explicitly through random sampling. 
This result demonstrates that our method of training sub-paths is more effective than training target sub-networks themselves.

\begin{wrapfigure}{r}{0.41\linewidth}
\vspace{-0.15in}
\includegraphics[width=0.92\linewidth]{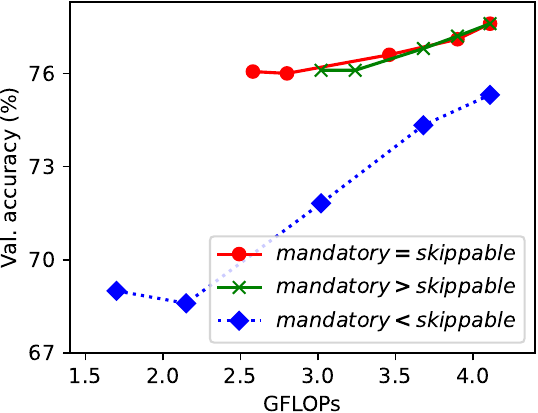}
  \caption{
    Pareto frontiers of three ResNet50-ADNs, each
    trained with varying ratios between mandatory and skippable sub-paths. 
    Total number of blocks remains unchanged.
 }
 \vspace{-0.3in}
\label{fig:skippableratio}
\end{wrapfigure}

\textbf{Lengths of Mandatory Sub-Paths}: 
In Figure \ref{fig:skippableratio}, we investigate
the impact of varying
the ratio of lengths between the mandatory and the skippable sub-paths.
(Details are in Table \ref{tab:varying-length-ratio}.) 
If mandatory sub-paths become shorter 
than skippable sub-paths, 
smaller sub-networks can be selected
since more layers can be skipped.
However, the result in Figure \ref{fig:skippableratio}
shows that this configuration (shown in blue line) significantly degrades performance of all sub-networks.
Since every sub-network shares parameters of 
mandatory sub-paths,
low inference capability of shallow 
mandatory sub-paths 
affects all sub-networks.
This implies that maintaining certain depths in mandatory sub-paths is crucial for effective inference.
Conversely, further increasing the length of mandatory sub-paths (shown in green line) does not further improve performance
and only reduces the range of depth adaptation.

%% file: conc.tex
\section{Conclusions}
\label{sec:conc}
We propose a practical approach to adaptive depth networks that can be applied to both CNNs and transformers with minimal training effort. 
We provide a general principle and a formal explanation 
on how depth adaptation can be achieved with minimal performance degradation. 
Under this principle, our approach can avoid typical exhaustive training of target sub-networks and instead focus on optimizing the sub-paths of the network to have specific properties.
At test time,   
these sub-paths can be connected  
in a combinatorial manner 
to construct sub-networks with various
accuracy-efficiency trade-offs from a single network.
Experimental results show 
that these sub-networks form a better 
Pareto frontier than non-adaptive 
baseline networks
and achieve actual inference acceleration.

%% file: appendix.tex
\section{Appendix: Detailed Settings and Evaluation Results}
\subsection{Detailed Architectures}
\label{sec:arch-detail}
\begin{table}[h]
    \begin{subtable}[h]{0.5\textwidth}
        \centering
        \begin{footnotesize}
        \renewcommand{\arraystretch}{0.9}
        \begin{tabular}{l | c  c | c}
              & number of  &  number of & \\
              & mandatory & skippable & total \\   
              & blocks & blocks & blocks \\    
        \hline \hline
        stage 1 & 2 &  1 & 3\\
        \hline
        stage 2 & 2 &  2 & 4\\
        \hline
        stage 3 & 3 &  3 & 6\\
        \hline
        stage 4 & 2 &  1 & 3\\
        \hline
       \end{tabular}
        \renewcommand{\arraystretch}{1.0}
       \end{footnotesize}
       \caption{ResNet50-ADN}
       \label{tab:week1}
    \end{subtable}
    \hfill
    \begin{subtable}[h]{0.5\textwidth}
        \centering
        \begin{footnotesize}
        \renewcommand{\arraystretch}{0.9}
        \begin{tabular}{l | c  c |  c}
              & number of  &  number of & \\
              & mandatory & skippable & total\\   
              & blocks & blocks & blocks \\    
        \hline \hline
        stage 1 & 2 &  1 & 3\\
        \hline
        stage 2 & 2 &  1 & 3\\
        \hline
        stage 3 & 2 &  1 & 3\\
        \hline
        stage 4 & 2 &  1 & 3\\
        \hline
       \end{tabular}
        \renewcommand{\arraystretch}{1.0}
       \end{footnotesize}
       \caption{Vit-b/16-ADN}
       \label{tab:week1}
    \end{subtable}
    
    \vspace{7pt}
    
    \begin{subtable}[h]{0.5\textwidth}
        \centering
        \begin{footnotesize}
        \renewcommand{\arraystretch}{0.9}
        \begin{tabular}{l | c  c | c}
              & number of  &  number of & \\
              & mandatory & skippable & total \\   
              & blocks & blocks & blocks\\    
        \hline \hline
        stage 1 &  1 & -  & 1\\
        \hline
        stage 2 & 1 &  1  & 2\\
        \hline
        stage 3 & 2 &  1 & 3\\
        \hline
        stage 4 & 2 &  2 & 4\\
        \hline
        stage 5 & 2 &  1 & 3\\
        \hline
        stage 6 & 2 &  1 & 3\\
        \hline
        stage 7 & 1 &  - & 1\\
        \hline
       \end{tabular}
       \renewcommand{\arraystretch}{1.0}
       \end{footnotesize}
       \caption{MbV2-ADN}
       \label{tab:week1}
    \end{subtable}
    \hfill
    \begin{subtable}[h]{0.5\textwidth}
        \centering
        \begin{footnotesize}
        \renewcommand{\arraystretch}{0.9}
        \begin{tabular}{l | c  c | c}
              & number of  &  number of &  \\
              & mandatory & skippable & total\\   
              & blocks & blocks & blocks \\    
        \hline \hline
        stage 1 & 1 &  1 & 2\\
        \hline
        stage 2 & 1 &  1 & 2\\
        \hline
        stage 3 & 3 &  3 & 6\\
        \hline
        stage 4 & 1 &  1 & 2\\
        \hline
       \end{tabular}
        \renewcommand{\arraystretch}{1.0}
       \end{footnotesize}
       \caption{Swin-T-ADN}
       \label{tab:week1}
    \end{subtable}

     \caption{Each stage of base models
     is evenly divided into two sub-paths;
     the first is mandatory and the other is skippable.
     Since ViT does not define hierarchical stages, 
     12 identical encoder blocks are divided into 4 stages,
     resembling other residual networks for vision tasks.
     }
     \label{tab:arch-details}
\end{table}

\subsection{Performance of Sub-Networks}
\label{sec:subnet-perf}
\begin{table}[h!]
    \begin{subtable}[h]{0.5\textwidth}
        \centering
        \begin{footnotesize}
        \begin{tabular}{c | c  c}
        sub-network & FLOPs (G) & Acc@1 (\%) \\
        \hline \hline
        FFFF & 4.11 & \textbf{77.6}\\
        \hline
        \textbf{T}FFF & 3.90 & \textbf{77.1}\\
        F\textbf{T}FF & 3.68 & 76.5\\
        FF\textbf{T}F & 3.46 & 75.6\\
        FFF\textbf{T} & 3.90 & 76.7\\
        \hline
        \textbf{TT}FF & 3.46 & \textbf{76.5}\\
        \textbf{T}F\textbf{T}F & 3.24 & 75.3\\
        \textbf{T}FF\textbf{T} & 3.68 & 76.4\\
        F\textbf{TT}F & 3.02 & 75.8\\
        F\textbf{T}F\textbf{T} & 3.46 & 75.9\\
        FF\textbf{TT} & 3.25 & 75.6\\
        \hline
        \textbf{TTT}F & 2.80 & 75.9\\
        \textbf{TT}F\textbf{T} & 3.24 & 75.8\\
        \textbf{T}F\textbf{TT} & 3.02 & 75.3\\
        F\textbf{TTT} & 2.80 & \textbf{76.0}\\
        \hline
        \textbf{TTTT} & 2.58 & \textbf{76.1} \\
        \hline
       \end{tabular}
       \end{footnotesize}
       \caption{ResNet50-ADN}
       \label{tab:week1}
    \end{subtable}
    \hfill
    \begin{subtable}[h]{0.5\textwidth}
        \centering
        \begin{footnotesize}
        \begin{tabular}{c | c  c}
        sub-network & FLOPs (G) & Acc@1 (\%) \\
        \hline \hline
        FFFF & 17.58 & \textbf{81.4}\\
        \hline
        \textbf{T}FFF & 16.20 & 81.1\\
        F\textbf{T}FF & 16.20 & 81.0\\
        FF\textbf{T}F & 16.20 & 80.6\\
        FFF\textbf{T} & 16.20 & \textbf{81.2}\\
        \hline
        \textbf{TT}FF & 14.67 & \textbf{80.9}\\
        \textbf{T}F\textbf{T}F & 14.67 & 80.4\\
        \textbf{T}FF\textbf{T} & 14.67 & 80.9\\
        F\textbf{TT}F & 14.67 & 80.5\\
        F\textbf{T}F\textbf{T} & 14.67 & 80.9\\
        FF\textbf{TT} & 14.67 & 80.6\\
        \hline
        \textbf{TTT}F & 13.21 & 80.5\\
        \textbf{TT}F\textbf{T} & 13.21 & \textbf{80.7}\\
        \textbf{T}F\textbf{TT} & 13.21 & 80.5\\
        F\textbf{TTT} & 13.21 & 80.6\\
        \hline
        \textbf{TTTT} & 11.76 & \textbf{80.6} \\
        \hline
       \end{tabular}
        \end{footnotesize}
        \caption{ViT-b/16-ADN}
        \label{tab:week2}
     \end{subtable}
     \caption{FLOPs and ImageNet validation accuracy of sub-networks.
     Only super-net (or, FFFF) and base-net (or, TTTT) are trained explicitly. 
     Sub-networks in the middle can be selected at test time without explicit training.
     The highest accuracy in each group is shown in bold.
     }
     \label{tab:subnet-perf}
\end{table}

\newpage

\subsection{Varying the Ratio of Sub-Path Lengths in ResNet50-ADN}

\begin{table}[h]
    \begin{subtable}[h]{0.5\textwidth}
        \centering
        \begin{footnotesize}
        \renewcommand{\arraystretch}{0.9}
        \begin{tabular}{l | c  c | c}
              & number of  &  number of & total \\
              & mandatory & skippable & blocks\\   
              & blocks & blocks & \\    
        \hline \hline
        stage 1 & 1 &  2 & 3\\
        \hline
        stage 2 & 1 &  3 & 4\\
        \hline
        stage 3 & 2 &  4 & 6\\
        \hline
        stage 4 & 1 &  2 & 3\\
        \hline
       \end{tabular}
        \renewcommand{\arraystretch}{1.0}
       \end{footnotesize}
       \caption{\# of mandatory $<$ \# of skippable}
       \label{tab:week1}
    \end{subtable}
    \hfill
        \begin{subtable}[h]{0.5\textwidth}
        \centering
        \begin{footnotesize}
        \renewcommand{\arraystretch}{0.9}
        \begin{tabular}{l | c  c | c}
              & number of  &  number of & total \\
              & mandatory & skippable & blocks\\   
              & blocks & blocks & \\    
        \hline \hline
        stage 1 & 2 &  1 & 3\\
        \hline
        stage 2 & 3 &  1 & 4\\
        \hline
        stage 3 & 4 &  2 & 6\\
        \hline
        stage 4 & 2 &  1 & 3\\
        \hline
       \end{tabular}
       \renewcommand{\arraystretch}{1.0}
       \end{footnotesize}
       \caption{\# of mandatory $>$ \# of skippable}
       \label{tab:week1}
    \end{subtable}
     \caption{
     The configurations of 
     ResNet50-ADNs
     with different proportions
     between mandatory and skippable sub-paths. 
     Total number of blocks at each stage remains unchanged.     
     }
     \label{tab:varying-length-ratio}
\end{table}

\subsection{Varying the Ratio of Sub-Path Lengths in Vit-b/16-ADN}
\begin{figure}[h]
  \centering
  \begin{footnotesize}
    \subfloat[\# of mandatory $<$ \# of skippable]{%
    \renewcommand{\arraystretch}{1.0}
    \begin{tabular}{l | c  c | c}
    & number of  &  number of & total \\
    & mandatory & skippable & blocks\\   
    & blocks & blocks & \\    
    \hline \hline
    stage 1 & 1 &  2 & 3\\
    \hline
    stage 2 & 1 &  2 & 3\\
    \hline
    stage 3 & 1 &  2 & 3\\
    \hline
    stage 4 & 1 &  2 & 3\\
    \hline
   \end{tabular}
    \renewcommand{\arraystretch}{1.0}
  }
  \end{footnotesize}
  \qquad
  \subfloat[Pareto frontiers of Vit-b/16-ADN]{
    \includegraphics[width=0.4\linewidth]{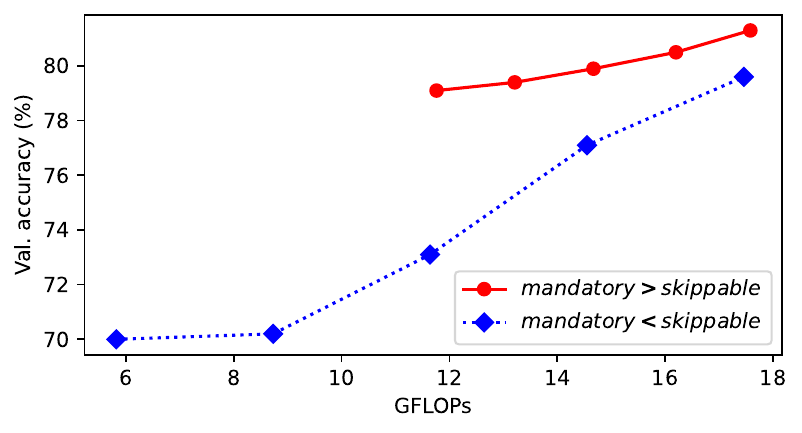}
  \label{subfig:black-hole}}%
  \caption{
  (a) The configuration of Vit-b/16-ADN with longer skippable sub-paths. (b) Pareto-frontier when different length ratios between 
  the mandatory and the skippable sub-paths are applied.}
  \label{fig:vit16-ratio}
\end{figure}

Each stage of Vit-b/16-ADN has 3 encoder blocks
and, by default, we select every last encoder block 
as a skippable sub-path.
Therefore every stage has two mandatory blocks and 1 
skipable blocks  as shown in Table \ref{tab:arch-details}-(b). 
Figure \ref{fig:vit16-ratio}-(a) shows different configuration
where every last two blocks of the stages 
become skippable. 
With this configuration in Figure \ref{fig:vit16-ratio}-(a), 
we can select much smaller sub-networks. 
For example, 
the smallest sub-network, or the base-net, of 
Vit-b/15-ADN has only 4 mandatory blocks and 
it requires 5.82 GFLOPs. 
However, the result in Figure \ref{fig:vit16-ratio}-(b)
shows that this configuration significantly degrades performance of all sub-networks.
As demonstrated with ResNet50-ADN in Figure \ref{fig:skippableratio},
low inference capability of shallow 
mandatory sub-paths affects all sub-networks.
This result again shows that maintaining certain depths in mandatory sub-paths is crucial for effective inference.

\section{Appendix: More Experiments and Analysis}
\subsection{Object Detection and Instance Segmentation}
\label{sec:exp_detection}

\renewcommand{\dblfloatpagefraction}{0.5}
\renewcommand{\arraystretch}{1.1}
\begin{table*}[tbhp!]
\centering
\begin{footnotesize}
\caption{Object detection and instance segmentation results on MS COCO dataset.}
\label{table:coco17box_mask}
  \begin{tabular}[tbhp]{lllcclcc}
  \hline 
   \multirow{2}{*}{Detector} & \multirow{2}{*}{Backbone} & \multirow{2}{*}{FLOPs} & \multicolumn{2}{c}{Individual Networks} & & \multicolumn{2}{c}{\textbf{ResNet50-ADN (ours)}} \\
  \cline{4-5} \cline{7-8} 
   &  & & Box AP & Mask AP & & \textbf{Box AP} & \textbf{Mask AP}\\
  \hline
  Faster-RCNN & ResNet50 & 207.07G & 36.4  &  &  & 37.8 & \\
  \cite{ren2017faster} & ResNet50-Base & 175.66G & 32.4 &  &  & 34.0 &\\
   \hline
   Mask-RCNN & ResNet50 & 260.14G &  37.2& 34.1 &  & 38.3 & 34.1\\
   \cite{he2017mask} & ResNet50-Base & 228.73G & 32.7 & 29.9  & & 34.1 & 31.2 \\
   \hline
   RetinaNet & ResNet50 & 151.54G & 36.4 &  &  & 37.4 & \\
  \cite{lin2017focal} & ResNet50-Base & 132.04G & 31.7 &  &   & 35.2\\
   \hline
  \end{tabular}
\end{footnotesize}
\end{table*}
\renewcommand{\arraystretch}{1.0}

In order to investigate the generalization ability of our approach,
we use MS COCO 2017 datasets 
on object detection and instance segmentation tasks using 
representative detectors.
We compare individual ResNet50 and our 
adaptive depth ResNet50-ADN
as backbone networks of the detectors.
For training of detectors, we use Algorithm \ref{alg:training} with slight adaptation.
For object detection, the intermediate features $\textbf{h}_{base}^s$
and $\textbf{h}_{super}^s (s=1..N_r)$
can be obtained directly from backbone network's feature pyramid networks (FPN) 
\cite{lin2017feature}, and, hence, 
a wrapper function is not required to extract intermediate features. 
All networks are trained on \texttt{train2017} for 12 epochs
from ImageNet pretrained weights, 
following the training settings
suggested in \cite{lin2017feature}. 
Table \ref{table:coco17box_mask} shows the results on \texttt{val2017}
containing 5000 images.
Our adaptive depth backbone networks
still outperform individual static backbone networks
in terms of COCO's standard metric AP. 

\subsection{Visual Analysis of Sub-Paths}
\label{sec:analysis-gradcam}

\begin{figure*}[htb!]
  \centering
    \begin{subfigure}[b]{0.77\linewidth}
        \includegraphics[width=\linewidth]{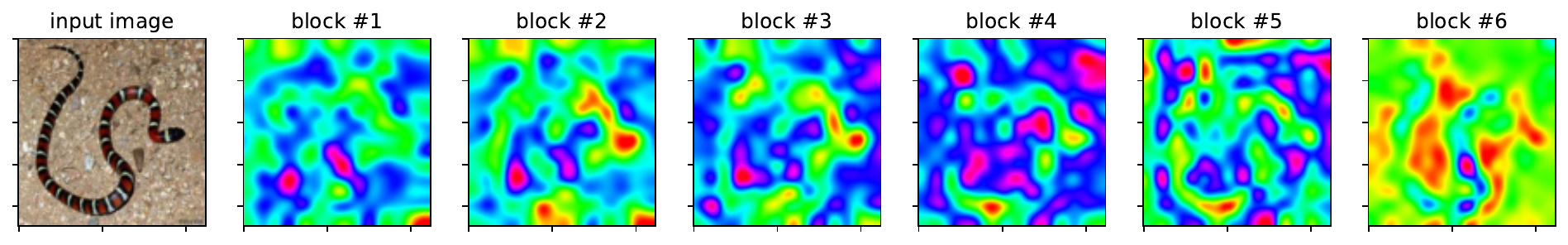}
    \caption{Original ResNet50}
  \end{subfigure}
  \begin{subfigure}[b]{0.77\linewidth}
        \includegraphics[width=\linewidth]{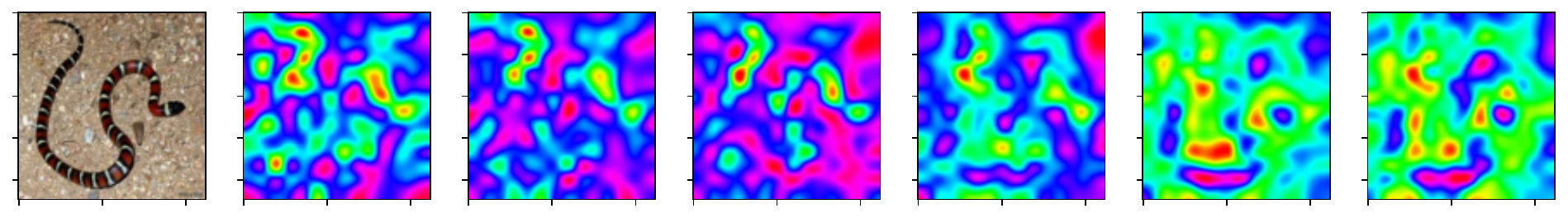}
    \caption{ResNet50-ADN (FFFF)}
  \end{subfigure}
  \begin{subfigure}[b]{0.77\linewidth}
    \includegraphics[width=\linewidth]{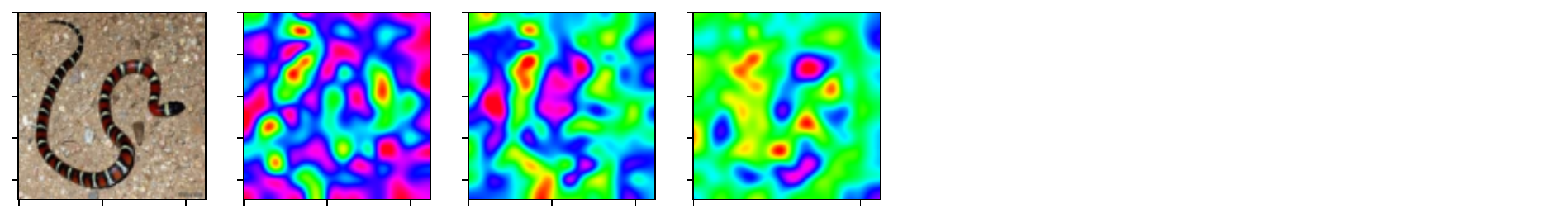}
    \caption{ResNet50-ADN (TTTT)}
  \end{subfigure}
  \caption{Class Activation Maps of the 3rd residual stages of ResNet50s.
  \textbf{(a)} Original ResNet50's activation regions change gradually 
  across all blocks.
  \textbf{(b)} In ResNet50-ADN (FFFF),
  the first 3 blocks have 
  extensive hot activation regions,  
  implying active learning of new level features.
  In contrast, 
  the skippable last 3 blocks have far less
  activation regions and they are gradually refined around the target. 
  \textbf{(c)} Even though parameters are shared,
  the activation map of base-net is very different from 
  super-net's
  since they use different batch normalization operators.
  } 
\label{fig:gradcam}
\end{figure*}

To investigate how our training method affects
feature representations in the mandatory and the skippable sub-paths,
we visualize the activation of 3rd residual stage of ResNet50-ADN
using Grad-CAM \cite{selvaraju2017grad}. 
The 3rd residual stage of ResNet50-ADN has 6 residual blocks 
and the last three blocks are skippable.
In Figure \ref{fig:gradcam}-(a), 
the activation regions of original ResNet50 changes gradually across all consecutive blocks.
In contrast, in Figure \ref{fig:gradcam}-(b),
ResNet50-ADN(FFFF), or super-net,
manifests very different activation regions in two sub-paths.
In the first three residual blocks, 
we can observe lots of hot activation regions in wide areas,
suggesting active learning of new level features.
In contrast, significantly less activation regions are found 
in the skippable last three blocks and 
they are gradually concentrated around the target object,
demonstrating the refinement of learned features. 
While ResNet50-ADN(TTTT), or base-net, shares the parameters 
with the super-net in the first 3 mandatory blocks,
their activation regions are very different from the 
super-net's.
This is because 
while the super-net and the base-net share 
parameters in the non-skippable mandatory blocks,
they use different batch normalization operators
in the mandatory sub-paths.
Further, in Figure \ref{fig:gradcam}-(c),
we can observe that the final activation map of the base-net 
is very similar to the super-net's final activation map 
in Figure \ref{fig:gradcam}-(b).
This implies that they have similar distributions
for the same inputs,
as suggested in Section \ref{sec:theory}.


